\documentclass[10pt,letterpaper]{article}
\usepackage{amsmath,amssymb,amsthm, bm}
\usepackage{nameref,hyperref}
\usepackage{lastpage,graphicx}
\usepackage{epstopdf}
\usepackage{float}
\usepackage{fullpage}

\begin{document}
\vspace*{0.2in}

\begin{flushleft}
{\Large
\textbf\newline{Fused Lasso Improves Accuracy of Co-occurrence Network Inference in Grouped Samples}
}
% Jeffrey Propster\textsuperscript{2},
% with the Lorem Ipsum Consortium\textsuperscript{\textpilcrow}
\\
\bigskip
% \textbf{2} Department of Biological Sciences, Northern Arizona University, Flagstaff, Arizona, United States of America
Daniel Agyapong\textsuperscript{1},
Briana H. Beatty\textsuperscript{2},
Peter G. Kennedy\textsuperscript{2},
Jane C. Marks\textsuperscript{3},
Toby D. Hocking\textsuperscript{4}
\\
\bigskip
\textbf{1} School of Informatics, Computing, and Cyber Systems, Northern Arizona University, Flagstaff, Arizona, United States of America
\\
\textbf{2} Department of Plant and Microbial Biology, University of Minnesota, Saint Paul, Minnesota, United States of America
\\
\textbf{3} Department of Biological Sciences, Northern Arizona University, Flagstaff, Arizona, United States of America
\\
\textbf{4} Département d'informatique, Université de Sherbrooke, Canada
\bigskip
\end{flushleft}

\begin{abstract}
Co-occurrence network inference algorithms have significantly advanced our understanding of microbiome communities. 
However, these algorithms typically analyze microbial associations within samples collected from a single environmental niche, often capturing only static snapshots rather than dynamic microbial processes. 
Previous studies have commonly grouped samples from different environmental niches together without fully considering how microbial communities adapt and reorganize their associations when faced with varying ecological conditions. 
Our study addresses this limitation by explicitly investigating both spatial and temporal dynamics of microbial communities. 

We analyzed publicly available microbiome abundance data from soil, aquatic, and host-associated environments, collected across multiple locations and time points, to evaluate algorithm performance in predicting microbial associations using our proposed Same-All Cross-validation (SAC) framework. 
SAC evaluates algorithms in two distinct scenarios: training and testing within the same environmental niche (Same), and training and testing on combined data from multiple environmental niches (All). 

To overcome the limitations of conventional algorithms, we propose fuser, an algorithm that, while not entirely new in machine learning, is novel for microbiome community network inference.
It retains subsample-specific signals while simultaneously sharing relevant information across environments during training.
Unlike standard approaches that infer a single generalized network from combined data, fuser generates distinct, environment-specific predictive networks. 
Our results demonstrate that fuser achieves comparable performance to existing algorithms such as glmnet when evaluated within homogeneous environments (Same), and notably reduces test error compared to baseline algorithms in cross-environment (All) scenarios, highlighting its robustness and suitability for modeling microbial associations across diverse spatio-temporal niches.

\end{abstract}

\section{Introduction}
Microbial communities represent complex ecological systems where numerous species interact in complex networks that influence its structure and function. 
Co-occurrence network inference algorithms have emerged as important tools for explaining these interactions, helping researchers understand the complex dynamics of microbiome associations in different environments~\cite{faust2012microbial,lima2015determinants}. 
Despite significant advances in microbial network analysis, there has been limited exploration of how microbial communities adapt and form associations across changing environments~\cite{baldo2017convergence}. 
Most existing research has focused on characterizing microbiome networks within single habitats or combined different environmental samples without preserving their ecological distinctions~\cite{weiss2016correlation}. 
This oversight obscures potentially important ecological patterns in how microbial associations vary across spatial and temporal niches~\cite{trosvik2015ecology}. 
As Ma et. al~\cite{ma2020earth} highlights that while individual networks have been extensively studied, the interconnection patterns across diverse environments are still yet to be unexplored. 
The dynamics of microbial communities across different environmental niches present particular challenges for predictive modeling for network inference.

Traditional algorithms often assume the same model parameters (e.g., regression weights, network edge strengths) apply equally whether working with combined data or with each dataset separately, neglecting their potential interdependencies and thus failing to capture distinct ecological dynamics of individual environments~\cite{knights2011bayesian,marcos2021applications}. 

This limitation becomes increasingly problematic when trying to predict microbiome associations across diverse environments, as these approaches ignore the ecological factors that shape community structures in different settings~\cite{costello2012application}. 
Recent methodological advances by~\cite{agyapong2025cross} have addressed some of these challenges through the introduction of novel cross-validation methods for training and testing co-occurrence network inference algorithms, providing a more robust framework for evaluating network quality for various algorithms.
Our study addresses this critical gap by investigating the spatial and temporal dynamics of microbial communities through the cross-validation analysis of sub-samples across diverse environments. 

Using publicly available microbiome datasets collected from diverse habitats and at different times, we evaluate how well algorithms predict microbial associations when samples are pooled across environments versus kept separate (Same-All Cross-validation, SAC)~\cite{thaiss2016microbiome,thompson2017}. 

This framework allows us to better understand how environmental ecological factors influence microbial associations and community assembly.
We evaluate the traditional glmnet and fuser, which is a novel algorithm in the context of microbiome network inference that preserves subsample-specific information during combined training, enabling it to generate distinct predictive networks for diverse environmental niches.
Unlike conventional approaches that apply uniform coefficients across combined datasets, fuser maintains contextual integrity while integrating data across environments~\cite{papoutsoglou2023machine}. 
Our approach builds upon recent cross-validation techniques introduced by~\cite{agyapong2025cross} for inference algorithm evaluation, extending them to preserve subsample-specific niches during network generation. 
The fuser algorithm not only achieves comparable performance to existing algorithms when trained and tested on homogeneous environmental subsamples but significantly outperforms baseline algorithms in cross-subsample predictions~\cite{moreno2021statistical}. 
Our findings have important implications for understanding the ecological drivers of microbiome community assembly and for developing more accurate predictive models across heterogeneous environmental niches.

\paragraph{Summary of contributions.}
We introduce Same-All Cross-validation (SAC), a two-regime protocol that contrasts within-habitat ("Same") and pooled-habitat ("All") microbiome abundance data prediction accuracy, providing the first rigorous benchmark for how well co-occurrence network algorithms generalize across environmental niches~\cite{agyapong2025cross,hocking2024soaksameotherallkfoldcrossvalidation}.  
Building on this framework, we adapt the fused-lasso algorithm, implemented in the open-source \texttt{fuser} package to microbiome data, yielding the first network-inference algorithm that shares information between habitats while preserving niche-specific edges~\cite{dondelinger2016hdlss,chen2010gflasso}.  
Benchmarks on public soil, aquatic and host-associated datasets show that \texttt{fuser} matches standard lasso (glmnet) performance in homogeneous settings yet lowers test error in cross-habitat prediction, mitigating both the false positives of fully independent models and the false negatives of fully pooled models~\cite{friedman2010glmnet}.  
Taxon-wise analyses further reveal species-specific optimal regularization, highlighting the complementarity of fused and unfused algorithms for different ecological roles.  
Together, SAC and \texttt{fuser} form a principled, data-driven toolbox for tracking how microbial interaction networks shift across space and time, enabling more reliable forecasts of microbiome community responses to environmental change~\cite{hallac2015network,shade2023rescue}.

\section{Microbiome Datasets}
We collected some publicly available grouped-sample microbiome datasets, each offering unique characteristics in terms of taxonomic diversity, sample size, and ecological niches. 
Table~\ref{tab:datasets} summarizes the key features of these datasets.
The \textbf{Human Microbiome Project (HMP)}~\cite{HMP2012} dataset represents one of the earliest comprehensive characterizations of the healthy human microbiome, surveying multiple body sites including oral, nasal, skin, gastrointestinal, and urogenital regions.
The \textbf{HMPv35}~\cite{Huttenhower2012} dataset features an expanded collection of 16S rRNA gene sequencing data processed through the QIIME pipeline with clustering at 97\% sequence similarity.
The \textbf{MovingPictures}~\cite{Caporaso2011} dataset stands as a landmark temporal study that tracked microbial communities from four body sites (tongue, palm, forehead and gut) in two individuals over a long period of time. 
This longitudinal sampling approach revealed remarkable insights into microbiome stability and responsiveness to environmental factors, capturing both day-to-day fluctuations and longer-term community dynamics that reflect individual lifestyle patterns.
The \textbf{qa10394}~\cite{qa10394} evaluates how common storage preservatives and temperature conditions affect the stability and reproducibility of human and canine faecal microbiome profiles, providing guidance for field-sampling workflows.
The \textbf{TwinsUK} dataset~\cite{Goodrich2016} comprises 16S profiles from 1,126 twin pairs and is explicitly designed to disentangle genetic and environmental contributions to community assembly, providing heritability estimates for hundreds of bacterial taxa.
The \textbf{necromass} dataset represents a focused study of bacterial and fungal communities associated with necromass decomposition under different experimental conditions. 
This dataset comprises sequencing data from four necromass treatment by time combinations: highly melanized or lowly melanized samples were collected at 1 month and 3 months of incubation in pine forest soils in Minnesota, USA, along with soil controls.
Unlike the large-scale human microbiome datasets, the necromass dataset is characterized by a substantially smaller taxonomic scope (36 taxa encompassing both bacterial and fungal genera) and sample size (69 samples), organized into 5 experimental subsets corresponding to the different treatment conditions. 
Notably, this dataset reflects a more narrow experimental environment and a focus on culturable microorganisms.
The dataset is particularly valuable for investigating bacterial-fungal interaction networks in soil decomposer communities.

All datasets exhibit high sparsity, which is typical for microbiome data due to the presence of rare taxa and the inherent complexity of microbial communities. 
This characteristic poses challenges for analytical algorithms, in the field of microbiome network inference and community structure analysis.

\begin{table}[h]
\centering
\caption{Publicly Available Grouped-Samples Microbiome Datasets}
\label{tab:datasets}
\begin{tabular}{lrrrr}
\hline
Dataset & No. of Taxa & No. of Samples & No. of Groups & Sparsity (\%) \\
\hline
HMPv13~\cite{HMP2012} & 5,830 & 3,285 & 71 & 98.16 \\
HMPv35~\cite{HMP2012} & 10,730 & 6,000 & 152 & 98.71 \\
MovingPictures~\cite{Caporaso2011} & 22,765 & 1,967 & 6 & 97.06 \\
qa10394~\cite{qa10394} & 9,719 & 1,418 & 16 & 94.28 \\
TwinsUK~\cite{TwinsUK} & 8,480 & 1,024 & 16 & 87.70 \\
necromass & 36 & 69 & 5 & 39.78 \\
\hline
\end{tabular}
\end{table}

\section{Methodology}
\subsection{Preprocessing and Data Preparation}
We preprocessed the microbiome datasets to ensure balanced representation across experimental groups and to prepare the data for downstream machine learning analyses. 
The preprocessing pipeline consisted of the following sequential steps:
We applied a log10 transformation with pseudocount addition (log10(x + 1)) to the raw OTU count data. 
This transformation helps stabilize variance across different abundance levels and reduces the influence of highly abundant taxa while preserving zero values~\cite{agyapong2025cross, weiss2017normalization}.
To ensure equal representation across experimental groups for cross-validation procedures, we standardized group sizes by calculating the mean group size and randomly subsampling an equal number of samples from each group. 
This approach prevents group size imbalances from biasing downstream analyses~\cite{Knights2011supervised, pasolli2016machine}.
We removed low-prevalence OTUs to reduce sparsity and potential noise in downstream models~\cite{papoutsoglou2023machine,Cao2021,Kurtz2015}.
Finally, we ensured that the resulting datasets contained equal numbers of samples per experimental group, with log-transformed OTU abundances ready for machine learning model training and evaluation.

\subsection{Same All Cross-validation (SAC)}
% TODO: Add a little background on CV and how it is used in microbiome network inference.
Cross-validation (CV) is a fundamental technique in machine learning and statistical modeling, used to assess the performance and generalizability of predictive models~\cite{stone1974cv}.
Later empirical studies confirmed that k-fold CV (typically k = 5 to 10) strikes a good balance between bias, variance and computational cost~\cite{kohavi1995study}.
In the field of microbiome network inference, CV is particularly important because it allows researchers to evaluate how well algorithms can generalize across varying environmental niches or temporal dynamics.
By systematically testing algorithms on different folds of microbiome data, researchers can identify which models are robust and effective in capturing the complex interactions within microbial communities~\cite{arlot2010survey, agyapong2025cross}.
This study introduces Same-All Cross-validation (SAC), a constrained variant of the Same/Other/All K-fold cross-validation (SOAK) framework originally proposed by Hocking et al. (2024) for measuring pattern similarity across data subsets~\cite{hocking2024soaksameotherallkfoldcrossvalidation}. 
Our Same-All Cross-validation (SAC) framework is designed to rigorously evaluate the performance and generalizability of microbiome network inference algorithms across diverse ecological habitats.
SAC builds upon the principles of traditional cross-validation but introduces two distinct validation scenarios:

1. \textbf{Same}: In this scenario, algorithms are trained and tested exclusively within the same habitat, assessing their performance within identical ecological niches. 
This approach allows for a focused evaluation of how well algorithms can capture the specific interactions and relationships present in a given environment.
Agyapong et al.~\cite{agyapong2025cross} utilized this CV scenario extensively on microbiome datasets, which is also used in traditional CV approaches. 
The core concept is a CV scenario where both training and testing data come from the same one specific habitat (e.g., only gut samples) or from multiple habitats mixed together and treated as a single unified dataset.
In both cases, the training and test sets are drawn from the same pool of environmental conditions, rather than testing the model's ability to generalize across different habitats.
This approach evaluates algorithm performance within a consistent habitat or within combined habitats treated as a unified environment.

2. \textbf{All}: In this scenario, algorithms are trained on data from all available habitats and then evaluated individually within each habitat.
If there is only one habitat, this scenario is equivalent to the "Same" scenario~\cite{hocking2024soaksameotherallkfoldcrossvalidation}. 
This explicitly tests the algorithms' ability to generalize across distinct ecological niches, providing insights into their robustness and applicability in diverse settings.

\begin{figure}[H]
\centering
\includegraphics[width=\textwidth]{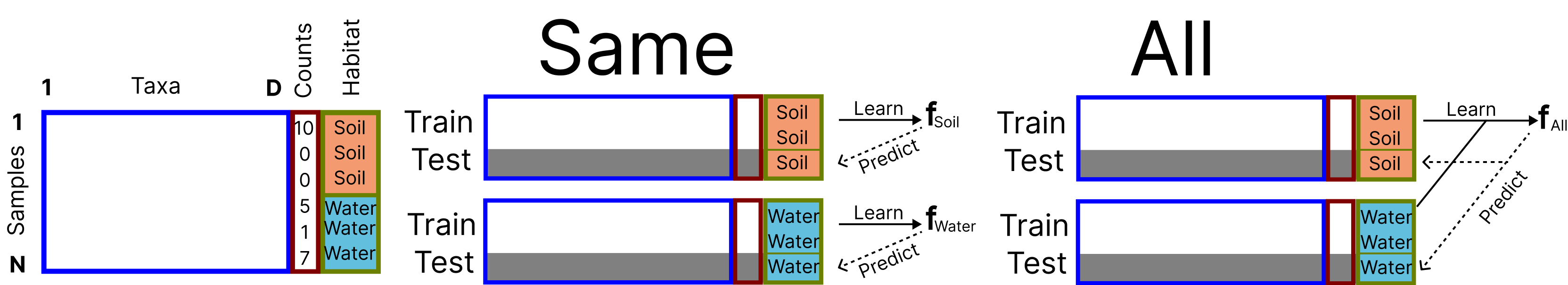}
\caption{Same All Cross-validation (SAC) for microbiome network inference across habitats.}
\label{fig:conceptual}
\end{figure}

Figure~\ref{fig:conceptual} illustrates our approach, depicting the underlying data structure and the two distinct validation scenarios: "Same" and "All".
The leftmost panel of the figure represents our microbiome dataset, comprising $N$ samples characterized by counts of $D$ microbial taxa. 
Each sample is categorized by its ecological niche or habitat type (e.g., soil, water). 
In the "Same" scenario, algorithms are trained and tested exclusively within the same habitat, assessing their performance within identical ecological niches. 
In the "All" scenario, algorithms are trained on data from all available habitats and then evaluated individually within each habitat, explicitly testing their ability to generalize across distinct ecological niches.
Our SAC strategy addresses the critical issue of overfitting, common in microbiome network inference, by testing algorithm performance under both scenarios. 
This allows identification of models capable of capturing ecological relationships that generalize beyond habitat-specific patterns, highlighting robustness and broad applicability. 
Additionally, this approach explores the potential for transfer learning in microbial ecology, suggesting that insights from integrated multi-habitat datasets can enhance network inference in individual ecological contexts.
By comparing the "Same" and "All" validation scenarios, we identify algorithms that not only excel within specific habitats but also effectively capture broader, habitat-independent ecological dynamics. 
This methodological advancement promises to deepen our understanding of microbial community structures across ecosystems, promoting more reliable and versatile network inference tools in microbiome research.

\paragraph{Taxon-wise Same-All Cross-validation (SAC).}
This technique employs the Leave One Out Cross-validation (LOOCV) strategy, where each taxon is treated as a separate prediction task.
This specific approach has been previously applied to compositional networks in CCLasso~\cite{fang2015cclasso}, now underpins microbe-disease predictors such as LRLSHMDA~\cite{wang2017lrlshmda}, and was recently formalised for co-occurrence benchmarking by Agyapong\emph{ et al.}~\cite{agyapong2025cross}.
Here, it is adapted to evaluate the performance of network inference algorithms across different habitats.
Let \(\mathcal{D}=\{(X_i,s_i,k_i)\}_{i=1}^{N}\) be a dataset of \(N\) microbiome samples,
where \(X_i\in\mathbb{N}^{D}\) is the count vector of \(D\) taxa,
\(s_i\in\{1,\dots,S\}\) is the habitat index,
and \(k_i\in\{1,\dots,K\}\) is the fixed \(K\)-fold label within that habitat.

\medskip
For each taxon \(d\in\{1,\dots,D\}\) we form a \textbf{taxon-specific prediction task} with response \(y_{i,d}=X_{i,d}\) and predictors
\(X_{i,-d}=(X_{i,1},\dots,X_{i,d-1},X_{i,d+1},\dots,X_{i,D})\).

\medskip
For every habitat \(\sigma\) and fold \(\kappa\) we define
\begin{align}
  \text{Test set:}\quad
  \mathcal{T}_{\sigma,\kappa} &=
     \bigl\{\,i \;\big|\; s_i=\sigma,\;k_i=\kappa\bigr\},
  \\[4pt]
  \text{Same train set:}\quad
  \mathcal{S}_{\sigma,\kappa} &=
     \bigl\{\,i \;\big|\; s_i=\sigma,\;k_i\neq\kappa\bigr\},
  \\[4pt]
  \text{All train set:}\quad
  \mathcal{A}_{\kappa} &=
     \bigl\{\,i \;\big|\; k_i\neq\kappa\bigr\}.
\end{align}

\medskip
A network-inference algorithm \(\operatorname{Alg}(\cdot)\) with
hyper-parameters~\(\theta\) is then fitted \emph{separately for each taxon}:
\begin{align}
  \hat{G}_{\sigma,\kappa,d}^{\mathrm{Same}}
      &=\operatorname{Alg}\!\bigl(\mathcal{S}_{\sigma,\kappa};\theta,d\bigr),
  &
  M_{\sigma,\kappa,d}^{\mathrm{Same}}
      &=\mathcal{M}\!\bigl(\hat{G}_{\sigma,\kappa,d}^{\mathrm{Same}},
                           \mathcal{T}_{\sigma,\kappa};d\bigr),
  \\[4pt]
  \hat{G}_{\sigma,\kappa,d}^{\mathrm{All}}
      &=\operatorname{Alg}\!\bigl(\mathcal{A}_{\kappa};\theta,d\bigr),
  &
  M_{\sigma,\kappa,d}^{\mathrm{All}}
      &=\mathcal{M}\!\bigl(\hat{G}_{\sigma,\kappa,d}^{\mathrm{All}},
                           \mathcal{T}_{\sigma,\kappa};d\bigr).
\end{align}

\medskip
Finally, we report performance averaged over \textbf{taxa, habitats, and folds}:
\[
  \overline{M}^{\mathrm{Same}}
     =\frac{1}{D S K}\sum_{d=1}^{D}\sum_{\sigma=1}^{S}\sum_{\kappa=1}^{K}
       M_{\sigma,\kappa,d}^{\mathrm{Same}},
  \qquad
  \overline{M}^{\mathrm{All}}
     =\frac{1}{D S K}\sum_{d=1}^{D}\sum_{\sigma=1}^{S}\sum_{\kappa=1}^{K}
       M_{\sigma,\kappa,d}^{\mathrm{All}}.
\]

\subsection{fuser: fused-lasso regularisation}
We implement fused-lasso regularisation using the subgroup-fusion approach proposed by Dondelinger and Mukherjee~\cite{dondelinger2016hdlss}.
Unlike the ordinary lasso, which only shrinks individual coefficients, the fused-lasso implemented in \texttt{fuser} adds an extra penalty on the pairwise differences between related coefficients, so it enforces both sparsity and smooth (or shared) effects across neighbouring features or sub-groups~\cite{tibshirani2005fused,dondelinger2016hdlss,fuser2025}.
We model each taxon $d\in\{1,\dots,D\}$ as the response and use the remaining taxa as predictors. 
For habitat
$s\in\{1,\dots,S\}$ and sample $i\in\{1,\dots,n_s\}$ let
\[
y_{i,d}^{(s)} = X_{i,-d}^{(s)} \,\beta_{d}^{(s)} + \varepsilon_{i,d}^{(s)},
\]
where $X_{i,-d}^{(s)}\in\mathbb{R}^{p}$ contains the counts of the
$p=D-1$ predictor taxa. We decompose each subgroup coefficient vector as
\[
\beta_{d}^{(s)} = \beta_{d}^{(0)} + u_{d}^{(s)},
\]
with a global part $\beta_{d}^{(0)}$ shared by all habitats and a
deviation $u_{d}^{(s)}$ specific to habitat $s$.
The parameters are estimated by minimising the taxon-specific objective
\[
\begin{aligned}
\mathcal{L}_{d}(\beta_{d}^{(0)},\{u_{d}^{(s)}\}) &= \sum_{s=1}^{S}\sum_{i=1}^{n_s}
  \bigl(y_{i,d}^{(s)} - X_{i,-d}^{(s)}(\beta_{d}^{(0)} + u_{d}^{(s)})\bigr)^{2} \\
&\quad + \lambda\!\Bigl(\lVert\beta_{d}^{(0)}\rVert_{1}
      + \sum_{s=1}^{S}\lVert u_{d}^{(s)}\rVert_{1}\Bigr) \\
&\quad + \gamma\!\sum_{s<t} w_{st}\,
      \lVert u_{d}^{(s)} - u_{d}^{(t)}\rVert_{1},
\end{aligned}
\]
where
% \begin{itemize}
% \item $\lambda > 0$ enforces sparsity of both global and subgroup coefficients (lasso term);
% \item $\gamma \geq 0$ enforces similarity across habitats via a \emph{fusion} term (information sharing);
% \item $w_{st}$ are optional non-negative weights encoding a prior graph of habitat similarity.
% \end{itemize}

\begin{itemize}
  \item $\lambda>0$ enforces sparsity of both global and subgroup coefficients (\emph{lasso} term)~\cite{tibshirani2005fused}.
  \item $\gamma\ge 0$ enforces similarity across habitats via a \emph{fusion} penalty that shrinks coefficient differences~\cite{tibshirani2005fused,dondelinger2016hdlss}.
  \item $w_{st}\!\ge\!0$ are optional weights encoding a prior graph of habitat similarity~\cite{hallac2015network,dondelinger2016hdlss}.
\end{itemize}
The optimisation is repeated for every taxon $d$; final network edges arise from the union of non-zero coefficients across taxa.
Cross-validated grids over $(\lambda,\gamma)$ are used to select the model that minimises prediction test error under the taxon-wise Same-All Cross-validation (SAC) technique.

\begin{figure}[h]
\centering
\includegraphics[width=\textwidth]{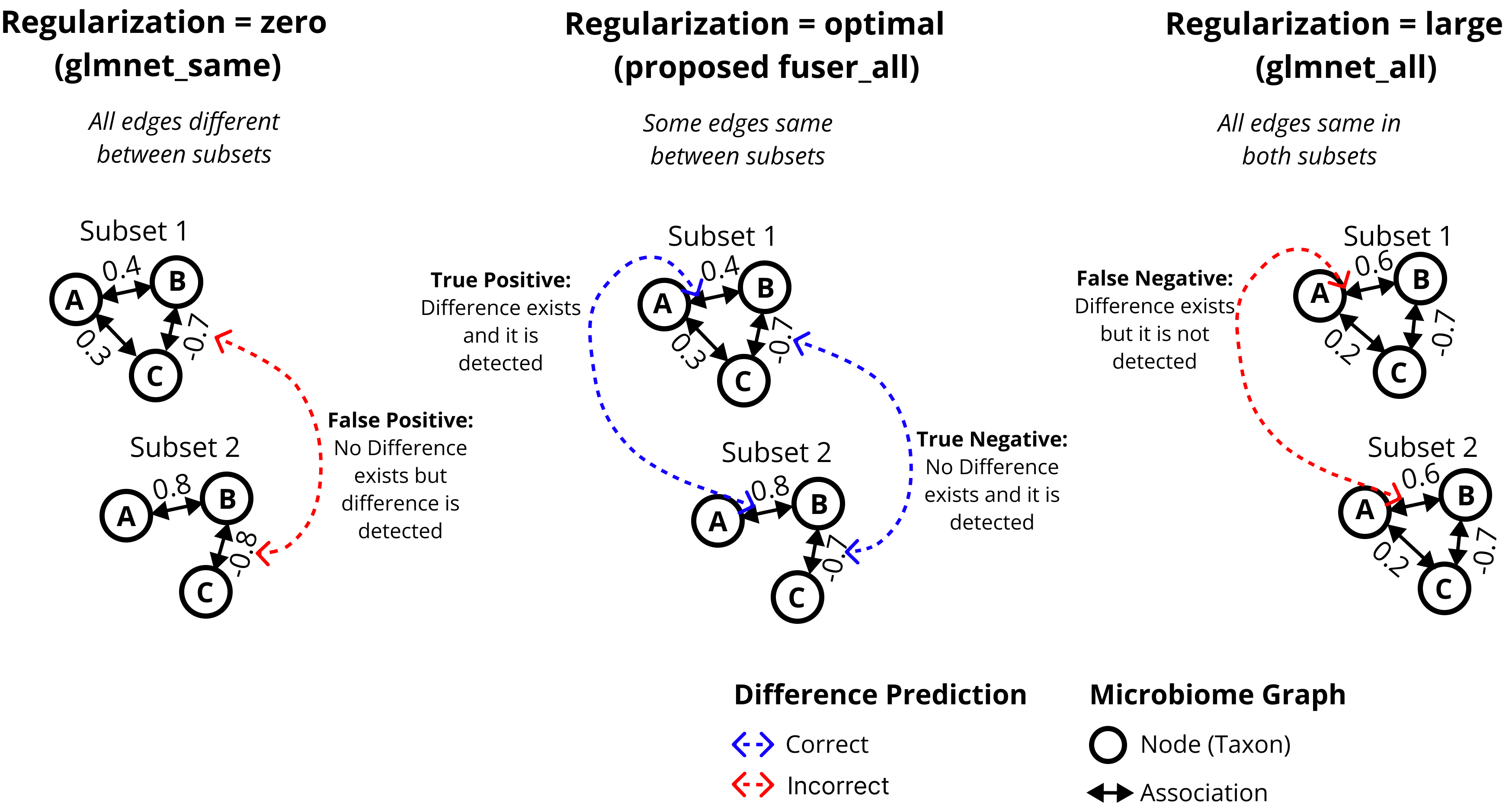}
\caption{Conceptual diagram of regularization for microbiome network inference across habitats.}
\label{fig:conceptual2}
\end{figure}

\subsubsection{Regularization Approaches}
This conceptual figure~\ref{fig:conceptual2} presents a critical comparison between three regularization approaches for inferring microbial association networks across different  environmental subsets.
The middle panel which corresponds to our proposed fuser\_all is the reference point that reveals what happens in actual microbial communities. 
Some microbial associations maintain consistent strength across different environmental subsets.
Genuine environmental adaptations occur in other associations (A-B varying from 0.4 to 0.8) that show biologically meaningful changes between environments.
This mixed pattern of stability and variability reflects fundamental ecological principles.
Each approach represents a fundamental trade-off between detecting genuine ecological differences and avoiding spurious connections~\cite{shade2012microbial}.
The three regularization approaches are illustrated in Figure~\ref{fig:conceptual2}
and summarized as follows:
\begin{enumerate}
    \item \textbf{Zero Regularization (glmnet\_same):}
    This approach treats all edges as potentially different between subsets, applying no penalty for variation because the algorithm is fit on each subset separately. 
    This allows complete independence of edge weights across environments because it treats environments as completely separate, losing statistical power and transfer learning.
    Hence, it results in complete flexibility to capture subset-specific interactions and high susceptibility to false positives (detecting non-existent differences).
    For example in the left figure, the B-C edge is flagged as different (-0.7 vs -0.8) while in the middle figure it is not (-0.7 vs -0.7). 
    In this case, the approach suffers from potential overfitting and so the variation is likely just sampling noise~\cite{friedman2010glmnet,overfit2024}.

    \item \textbf{Large Regularization (glmnet\_all):}
    This approach forces all edge weights to be identical across environments.
    This is because the algorithm is fit on all the subset at the same time.
    It collapses environment-specific variation into averaged values (A-B forced to 0.6 in both subsets).
    It eliminates noise but also eliminates genuine ecological differences.
    This results in false negatives by missing real biological variation (the A-B connection difference)~\cite{friedman2010glmnet}.

    \item \textbf{Optimal Regularization (fuser\_all):}
    The novel approach balances between enforcing similarity and allowing meaningful differences.
    It selectively preserves edges that truly differ between environments (A-B: 0.4 vs 0.8) and maintains consistent weights for genuinely stable associations (B-C remains -0.7 in both subsets).
    It achieves both true positives (detecting real differences) and true negatives (correctly identifying stable edges).
    The regularization is done by tuning the regularization parameters, lambda and gamma to achieve the best balance between preserving genuine ecological differences and suppressing noise.
    The key innovation lies in how fuser applies regularization. 
    Rather than applying uniform penalties across all edges, it selectively constrains variation based on evidence from the data itself~\cite{tibshirani2005fused,dondelinger2016hdlss,fuser2025}.
\end{enumerate}

\section{Results}
\subsection{Is it better to combine environmental subsets?}
\begin{figure}[H]
    \centering
    \includegraphics[width=0.85\textwidth]{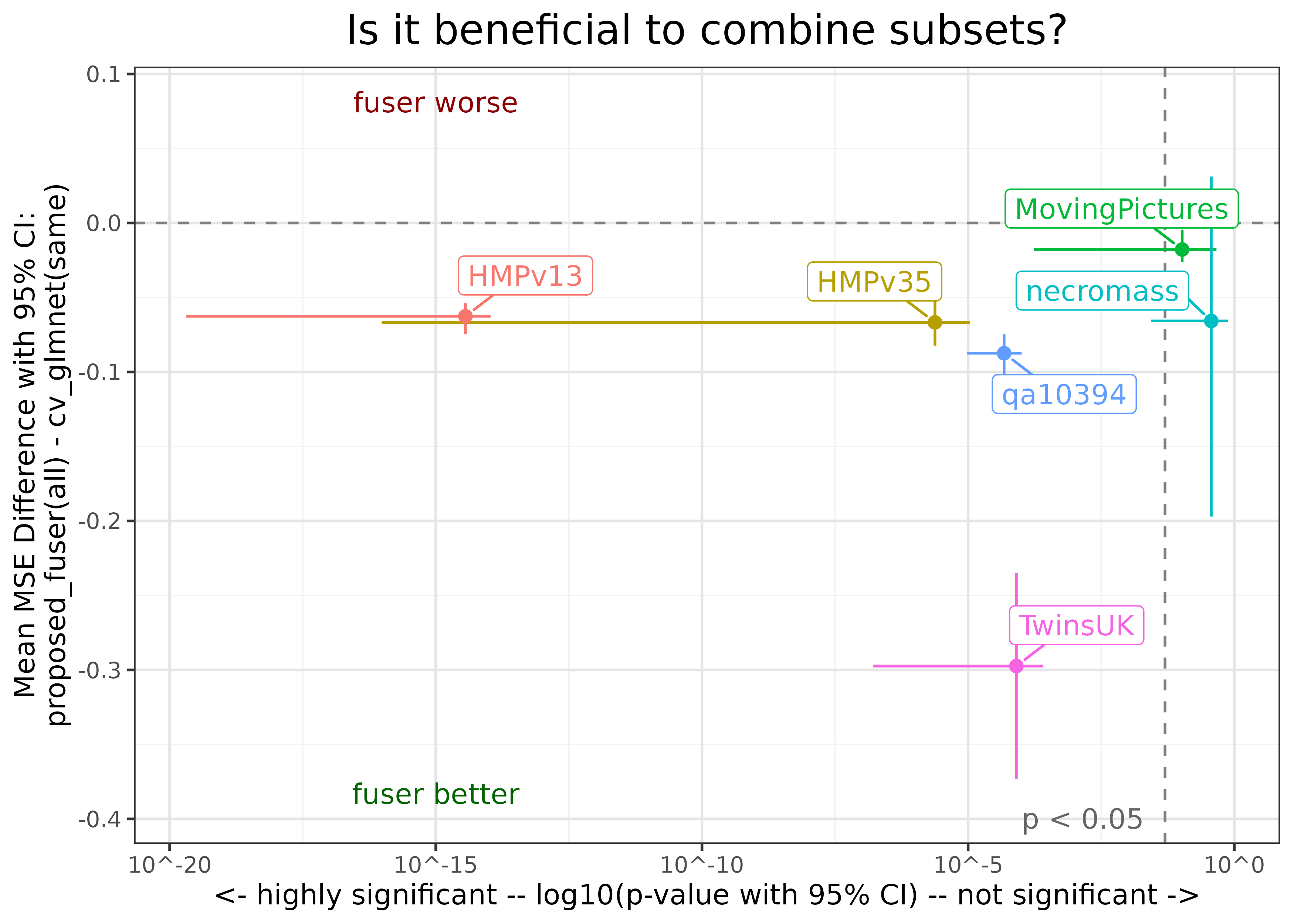}
    \caption{\textbf{Performance comparison between \texttt{fuser} using "all" available subsets and \texttt{cv\_glmnet} using "same" subsets only.} 
    The y-axis shows mean MSE difference with 95\% confidence intervals between \texttt{fuser(all)} and \texttt{cv\_glmnet(same)}. 
    Points below zero indicate \texttt{fuser} performs better when combining subsets compared to \texttt{cv\_glmnet} trained on individual subsets, while points above zero indicate worse performance. 
    The x-axis shows $\log_{10}$(p-value with 95\% CI), with the vertical dashed line at $p = 0.05$ representing the significance threshold.
    All datasets show improved performance when \texttt{fuser} combines environmental subsets.
    }
    \label{fig:fuserall_glmnetsame_diff}
\end{figure}

Figure~\ref{fig:fuserall_glmnetsame_diff} compares the performance of our \texttt{proposed\_fuser} algorithm~\cite{fuser2025,dondelinger2016hdlss} against the conventional \texttt{cv\_glmnet} algorithm~\cite{friedman2010glmnet} when both algorithms use all available subsets for training and testing.
This represents the "All" scenario for the \texttt{proposed\_fuser} algorithm and the "Same" scenario for the \texttt{cv\_glmnet} algorithm in our Same-All Cross-validation (SAC) framework~\cite{agyapong2025cross, hocking2024soaksameotherallkfoldcrossvalidation}.
The results demonstrate that \texttt{fuser} consistently outperforms \texttt{cv\_glmnet} across all datasets, with the magnitude and statistical significance varying by dataset characteristics.
The most substantial improvement occurs in the \textit{TwinsUK} dataset~\cite{TwinsUK}, where \texttt{fuser} achieves a mean squared error (MSE) reduction of approximately 0.30 ($p < 0.05$), representing the largest performance gain observed. 
For datasets \textit{qa10394}~\cite{qa10394}, \textit{HMPv13}~\cite{HMP2012}, and \textit{HMPv35}~\cite{HMP2012}, \texttt{fuser} shows statistically significant improvements ($p < 0.05$) with MSE reductions ranging from 0.05 to 0.1.
These significant improvements suggest that these datasets contain meaningful environmental heterogeneity that \texttt{fuser}'s fusion penalty can effectively leverage to improve predictive accuracy.
The other datasets (\textit{MovingPictures}~\cite{Caporaso2011} and \textit{necromass}) show modest improvements favoring \texttt{fuser} (MSE reductions of approximately 0.02-0.07), but these differences do not reach statistical significance.
This pattern suggests that while \texttt{fuser} maintains competitive performance, the environmental structure in these datasets may be less pronounced making the fusion penalty provide only subtle benefits.
Overall, the results support the theoretical framework that environmental heterogeneity in microbial associations can be effectively modeled through fusion penalties, with \texttt{fuser} providing consistent improvements across diverse microbiome datasets while achieving statistical significance in datasets with more pronounced environmental variability~\cite{fuser2025,dondelinger2016hdlss}.

\subsection{Is the proposed fuser better than the conventional cv\_glmnet when both use all subsets?}
\begin{figure}[H]
    \centering
    \includegraphics[width=0.85\textwidth]{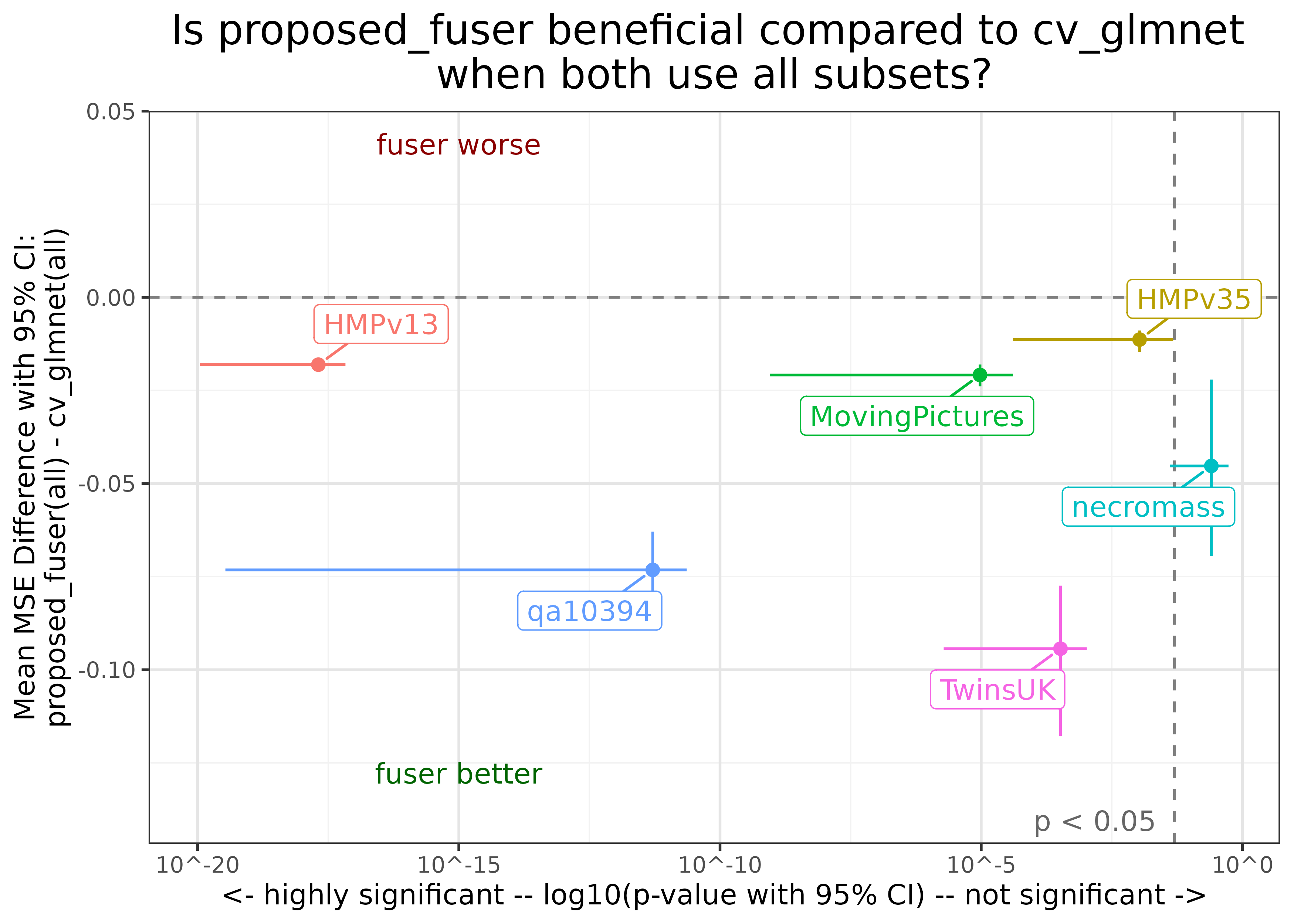}
    \caption{\textbf{Performance comparison between \textsc{fuser} and \texttt{cv\_glmnet} when both methods utilize all available subsets.} 
    The y-axis shows mean MSE difference with 95\% confidence intervals between \textsc{fuser(all)} and \texttt{cv\_glmnet(all)}. 
    Negative values (below zero) indicate \texttt{fuser} performs better, while positive values indicate \texttt{cv\_glmnet} performs better. 
    The x-axis shows $\log_{10}$(p-value with 95\% CI), with the vertical dashed line at $p = 0.05$ representing the significance threshold.}
    \label{fig:fuserall_glmnetall_diff}
\end{figure}

Figure~\ref{fig:fuserall_glmnetall_diff} addresses a fundamental methodological question in microbiome network analysis: when both algorithms have access to the same comprehensive dataset spanning multiple environmental subsets, does \texttt{fuser}'s fusion-based regularization provide superior predictive accuracy compared to conventional lasso regularization?
The results demonstrate that \texttt{fuser} consistently outperforms \texttt{cv\_glmnet} across all datasets when both methods utilize all available subsets.
All six datasets show improved performance favoring \texttt{fuser}, with mean squared error reductions ranging from near-zero to approximately 0.10. 
Three datasets (\textit{TwinsUK}~\cite{TwinsUK}, \textit{qa10394}~\cite{qa10394}, and \textit{HMPv13}~\cite{HMP2012}) achieve statistically significant improvements ($p < 0.05$), with \textit{TwinsUK}~\cite{TwinsUK} showing the most substantial performance gain.
The \textit{necromass} dataset shows a notable improvement that approaches but does not reach statistical significance.
The consistent direction of improvement across all datasets suggests that \texttt{fuser}'s fusion penalty provides a more appropriate regularization framework for microbiome data characterized by environmental heterogeneity.
Even when both algorithms have access to identical training data, \texttt{fuser}'s ability to model subset-specific patterns while sharing information across environmental niches appears to capture the underlying ecological structure more effectively than conventional lasso regularization.
These findings demonstrate that the theoretical advantages of fusion-based regularization translate into practical improvements in predictive accuracy, supporting the adoption of \texttt{fuser} for microbiome network inference tasks involving environmentally heterogeneous data~\cite{fuser2025,dondelinger2016hdlss}. 

\subsection{Taxa-specific performance reveals algorithm complementarity}
\begin{figure}[t]
    \centering
    \includegraphics[width=0.85\textwidth]{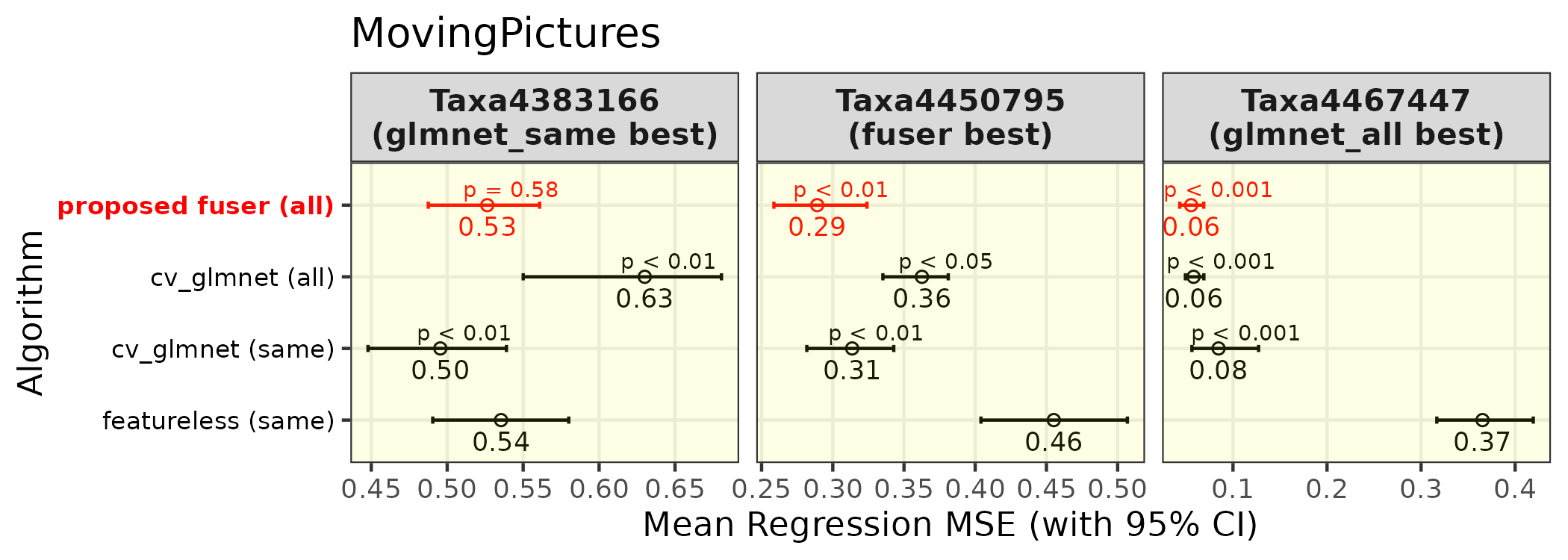}
    \caption{\textbf{Taxa-specific performance comparison across regularization methods in MovingPictures dataset.} 
    Each panel shows mean regression MSE with 95\% confidence intervals for a representative taxon, comparing \texttt{proposed\_fuser(all)}, \texttt{cv\_glmnet(all)}, \texttt{cv\_glmnet(same)}, and featureless baseline algorithms. 
    Different taxa exhibit distinct optimal methods: Taxa4383166 favors subset-specific modeling, Taxa4450795 benefits from fusion regularization, and Taxa4467447 performs best with standard regularization on combined data. 
    P-values indicate statistical significance of performance differences.}
    \label{fig:taxa_performance_movingpictures}
\end{figure}

While analyses on the dataset level demonstrate overall trends, examining individual taxon reveals substantial heterogeneity in optimal regularization strategies across microbial species. 
Figure~\ref{fig:taxa_performance_movingpictures} illustrates this pattern using representative taxa from the \textit{MovingPictures} dataset, where different taxa exhibit distinct preferences for regularization approaches.
The results demonstrate that no single algorithm universally outperforms others across all taxa within a dataset. 
For example, Taxa4383166 achieves lowest prediction error with \texttt{cv\_glmnet(same)} ($p < 0.01$), suggesting this taxon benefits from subset-specific modeling without information sharing.
In contrast, Taxa4450795 shows significantly improved performance with \texttt{proposed\_fuser(all)} ($p < 0.01$), indicating this taxon's associations are enhanced by fusion-based regularization across environmental niches. 
Taxa4467447 performs optimally under \texttt{cv\_glmnet(all)}, suggesting its associations are consistent across environments and benefit from increased sample size without specialized regularization.
Taxa with conserved ecological roles across environments may benefit from standard regularization with larger sample sizes, while environmentally responsive taxa may require fusion-based approaches to capture subset-specific association patterns~\cite{fuser2025,dondelinger2016hdlss}.  

\section{Discussion and Conclusion}
\subsection{Necromass microbial community difference networks}
\begin{figure}[t]
\centering
\includegraphics[width=\textwidth]{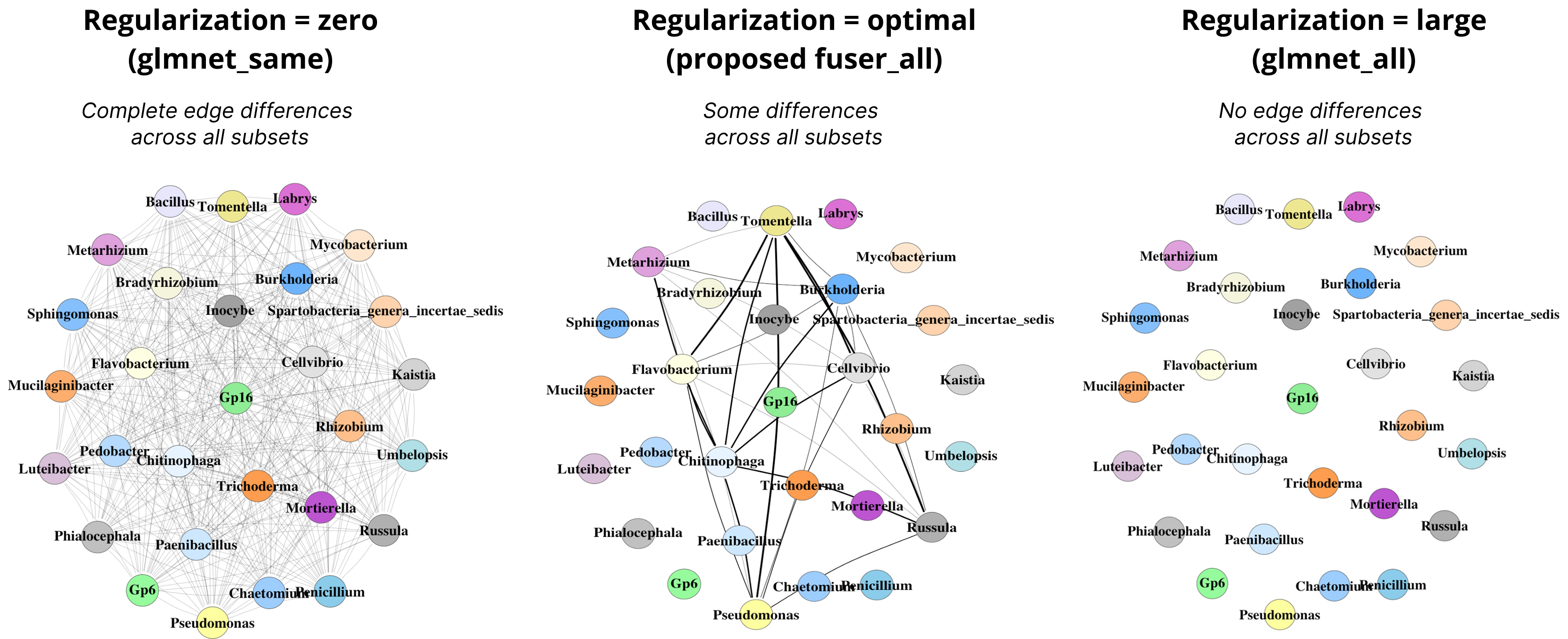}
\caption{\textbf{Impact of regularization on detected edge weight differences in the necromass microbial community.} 
Node positions are preserved across panels, with each node representing a taxon.
Edges indicate detected differences in association strength between subsets, with edge weights reflecting the magnitude of difference.
No edges indicate consistent associations across subsets.
}
\label{fig:necromass_networks}
\end{figure}

In this analysis, we focused on comparing two distinct treatment groups from our experimental design: AllSoilM1M3 (soil samples from both Month 1 and Month 3 timepoints) and LowMelM1 (low melanization necromass samples from Month 1 timepoint). This subset was selected to examine the fundamental differences between soil and necromass microbial communities while controlling for melanization level and minimizing temporal variation effects.
Figure~\ref{fig:necromass_networks} illustrates how different regularization strategies affect the detection of edge weight differences across environmental subsets in the necromass microbial real data. 
In these difference networks, an edge between taxa indicates a difference in association strength between subsets, while absence of an edge signifies consistency across environments.

Zero regularization (\texttt{glmnet\_same}) produces a dense, fully connected difference network. 
This approach forces all microbial associations to be different and independent across environmental subsets, resulting in a difference network that fails to capture genuine ecological differences.
For example, the \textit{Sphingomonas} and \textit{Burkholderia} association is forced to be different across the AllSoilM1M3 and LowMelM1 subsets, even though these taxa likely maintain consistent ecological relationships regardless of habitat type. 
This artificial differentiation obscures genuine ecological patterns by treating all microbial interactions as environment-specific when many core associations may be conserved across soil and necromass communities.

Strong regularization (\texttt{glmnet\_all}) treats all microbial associations as identical across subsets, resulting in a fully sparse difference network where no edges are detected. 
This indicates that the algorithm fails to identify any differences in microbial association strengths, even when they may exist.
Even ecologically important taxa such as \textit{Sphingomonas}, \textit{Burkholderia}, and \textit{Flavobacterium} show no differential connectivity between soil and low melanization necromass environments, suggesting the algorithm fails to detect biologically meaningful habitat-specific associations. 
This excessive regularization essentially assumes perfect ecological equivalence between soil and necromass communities, which contradicts known biological differences in nutrient availability, pH, and microbial succession patterns between these distinct habitats.

Our proposed \texttt{fuser\_all} algorithm achieves optimal balance in difference detection. 
Optimal regularization selectively identifies statistically supported edge weight differences. 
The intermediate network density reveals which microbial associations genuinely vary across environments while identifying those that remain consistent. 
The selective regularization allows truly environment-invariant relationships to be recognized (absence of edges) while preserving statistically supported differences (presence of edges).
This selective approach reveals a biologically plausible difference network where core soil-necromass associations involving taxon like \textit{Bradyrhizobium} and \textit{Burkholderia} show consistent patterns across both habitats (absence of difference edges), while associations involving taxa such as \textit{Chitinophaga} and \textit{Paenibacillus} display significant habitat-specific variation (presence of difference edges). 
The detected differences align with known ecological roles:  \textit{Chitinophaga}'s the enhanced activity of \textit{Chitinophaga} in decomposing environments and the variable environmental responses of \textit{Paenibacillus}.
The necromass difference networks demonstrate that \texttt{fuser\_all} captures the critical balance between over-detection of differences and under-detection of differences. 
This optimal difference detection ideally should translate directly to superior predictive performance, with the cross-validation tests. 
However, there are times that even with the proposed \texttt{fuser} algorithm has its predictive performance not better than the traditional \texttt{glmnet} even when all the subsets are used in training.
This suggests that while \texttt{fuser} provides a more nuanced understanding of ecological differences, it may not always outperform simpler models in terms of predictive accuracy, particularly when the ecological differences are subtle or when the data is highly sparse~\cite{fuser2025,dondelinger2016hdlss,friedman2010glmnet}.

\subsection{Conclusion}
Previous algorithms to microbial network inference across environments have either treated subsets as entirely independent or forced complete uniformity~\cite{matchado2021network, Kurtz2015}. 
Our findings demonstrate that both extremes result in sub-optimal inference, with excess false positives or false negatives in difference detection~\cite{fang2017gcoda,hebiri2011smoothlasso}.
The \texttt{fuser} algorithm bridges this methodological gap, offering a principled approach to selectively regularize edge-weight differences based on statistical evidence~\cite{sankaran2019multitable,stevens2021gwr,dondelinger2016hdlss}.
This allows \texttt{fuser} to achieve superior predictive performance while preserving biologically meaningful ecological patterns~\cite{sankaran2019multitable}.
While our approach significantly improves cross-environment inference, several limitations should be acknowledged~\cite{dohlman2019interactome}.
First, the optimal regularization parameters remain data-driven, requiring careful cross-validation for each application which is computationally intensive~\cite{hocking2024soaksameotherallkfoldcrossvalidation, agyapong2025cross}.
Second, while \texttt{fuser} captures environmental subset-specific differences, it may not fully account for phylogenetic relationships or functional potentials of taxa, which could further refine ecological interpretations~\cite{liu2024phylo,kunin2005phylogenetic,Douglas2020}.
Finally, the algorithm's performance is contingent on the quality and representativeness of the input abundance data, which can vary widely across microbiome studies~\cite{Sinha2017,McLaren2019,Gibbons2018}.
Despite these limitations, our results demonstrate that \texttt{fuser} provides a robust framework for microbial network inference across environmental subsets, effectively balancing the trade-offs between overfitting and underfitting.
Future work should explore the application of this SAC technique to incorporate phylogenetic information and metabolic potentials to further refine our understanding of when and why microbial associations differ across environments~\cite{liu2024phylo,dohlman2019interactome}.

\section*{Declarations}
\subsection*{Ethics approval and consent to participate}
Not applicable

\subsection*{Consent for publication}
Not applicable

\subsection*{Availability of data and materials}
The data and code used in this study are publicly available at \\ \url{https://github.com/EngineerDanny/necromass}.

\subsection*{Competing interests}
The authors declare that they have no competing interests.

\subsection*{Funding}
This work was funded by the National Science Foundation grant \#2125088 from Rules of Life Program.

\subsection*{Authors' contributions}
DA conceived the study, implemented the software, performed the analyses, and drafted the manuscript.  
BHB generated the necromass dataset.  
PGK revised the manuscript for biological relevance.
TDH advised on methodology, algorithm design, and manuscript revisions.  
All authors read and approved the final manuscript.  

\subsection*{Acknowledgments}
The authors would like to thank F. Maillard for oversight in generating the necromass dataset as well as the Microbes Persist Soil Microbiome SFA Award SCW1632 from the U.S. Department of Energy, Office of Biological and Environmental Research, Genomic Science Program, which facilitated this research collaboration.

\subsection*{Authors' information}
Not applicable

\bibliographystyle{plain}
\bibliography{references}

\end{document}